\documentclass[10pt,twocolumn,letterpaper]{article}

\usepackage[pagenumbers]{cvpr} 

\usepackage{graphicx}
\usepackage{amsmath}
\usepackage{amssymb}
\usepackage{booktabs}
\usepackage{float}
\usepackage[hyphens]{url}
\urldef{\footurl}\url{https://github.com/stepankonev/waymo-motion-prediction-chalenge-2022-multipath-plus-plus}

\usepackage[pagebackref,breaklinks,colorlinks]{hyperref}

\usepackage[capitalize]{cleveref}
\crefname{section}{Sec.}{Secs.}
\Crefname{section}{Section}{Sections}
\Crefname{table}{Table}{Tables}
\crefname{table}{Tab.}{Tabs.}

\def\cvprPaperID{*****} 
\def\confName{CVPR}
\def\confYear{2022}

\begin{document}

\title{MPA: MultiPath++ Based Architecture for Motion Prediction}

\author{Stepan Konev\\
{\tt\small stevenkonev@gmail.com}
}
\maketitle

\begin{abstract}
   Autonomous driving technology is developing rapidly and nowadays first autonomous rides are being provided in city areas. This requires the highest standards for the safety and reliability of the technology. Motion prediction part of the general self-driving pipeline plays a crucial role in providing these qualities. In this work we present one of the solutions for Waymo Motion Prediction Challenge 2022 based on MultiPath++ \cite{mppp} ranked the 3rd as of May, 26 2022. Our source code is publicly available on GitHub\footnote[1]{\footurl \label{ftn:source}}.
\end{abstract}

\section{Introduction}
\label{sec:intro}

The most popular approach for creating an autonomous driving technology consists of multiple steps: receiving data from sensors, solving perception problem to recognize the surrounding objects, localization, motion prediction and finally motion planning. In this paper we focus solely on the motion prediction problem, considering the relative locations of other agents and road lines as ground truth. Motion prediction problem has been approached in multiple works however it remains a complex and challenging problem so far. One of the key difficulties is a natural uncertainty of other agents' behavior. In this paper we propose a solution based on MultiPath++ \cite{mppp} which efficiently scores the 3rd on Waymo Motion Prediction Challenge 2022 and significantly outperforms our previous solution based on Convolutional Neural Network.

\section{Related work}
Motion prediction task is a challenging problem that attracts a lot of attention from researchers. There are two main approaches for this task: based on CNN \cite{lee2017desire,cui2019multimodal,chai2019multipath,hong2019rules,phan2020covernet,kawasaki2021multimodal} and GNN \cite{casas2019spatially,gao2020vectornet,nms_dense_tnt}. For CNN the input data is represented as a dense tensor where the channel dimension is usually referred to as a discrete-time dimension. The GNN approach takes graph data as input where the road lines and previous agents' positions are represented as nodes within polylines. More details about our specific representation are in section \ref{input_data}. However CNN approach seems comfortable for its simplicity because CNNs are greatly explored and stable due to the regular structure of the input data it tends to lose its popularity in favor of the GNN approach.

\section{Method}
In this section we provide the description of the input data structure, its preprocessing, model architecture and training strategy followed by a postprocessing description.
\subsection{Input data}
\label{input_data}
For training our model we first did some preprocessing for agents of interest. Since our model is based on MultiPath++ \cite{mppp} we follow similar process of data input data preparation as in MultiPath++ \cite{mppp}. A usual approach is to first transform the frame into the canonical coordinate system where the agent we make prediction for is always located in the same position with the same heading at the moment we make prediction for. This step helps us to eliminate redundant symmetries. Road graph data was represented as in MultiPath++ \cite{mppp}. For the target agent and other agents that surround the target agent for each timestamp in history (past + current) we computed $x,y$ coordinates, heading, velocity in the mentioned canonical coordinate system and the validity boolean flag. We cached this precomputed data along with road graph data for faster training. The size of the dataset and corresponding splits remain the same as described in our previous work MotionCNN \cite{wmpc2021}.
\subsection{Model architecture}
\begin{figure*}[t!]
\begin{center}
\includegraphics[width=1.0\linewidth]{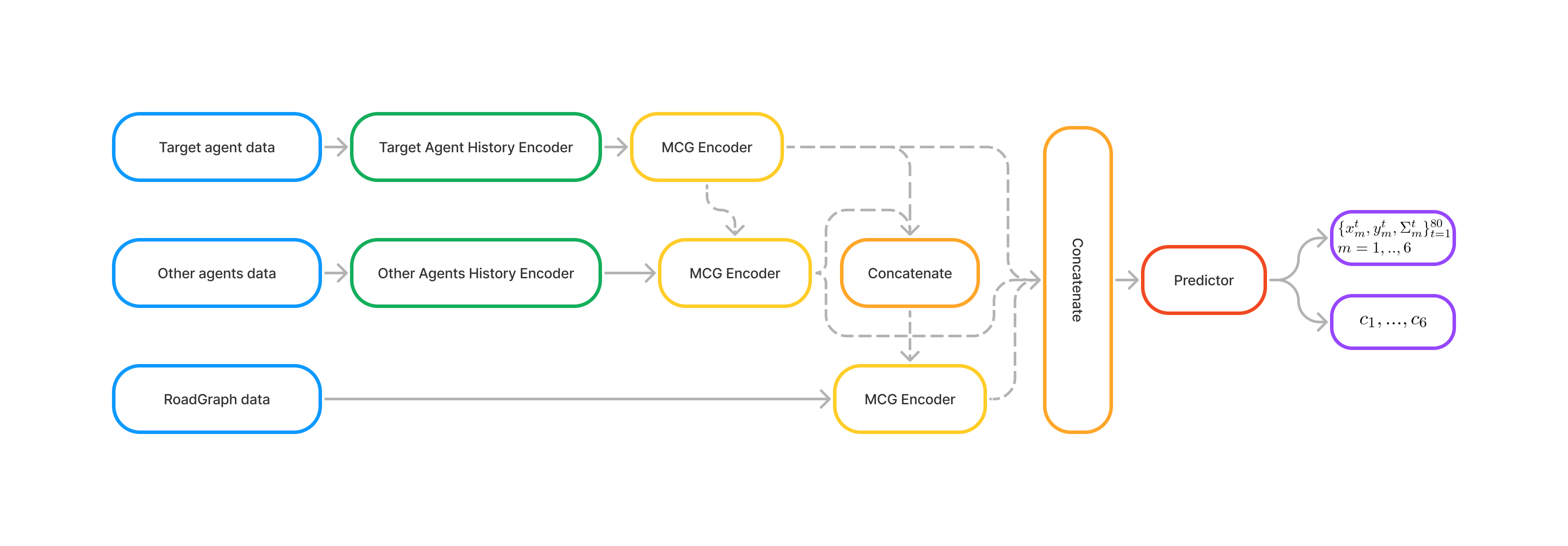}
\end{center}
   \caption{General overview of the architecture of out model based on MultiPath++ \cite{mppp}}
\label{fig:short}
\label{fig:pipeline}
\end{figure*}

\begin{figure}[t!]
\begin{center}
\includegraphics[width=1.0\linewidth]{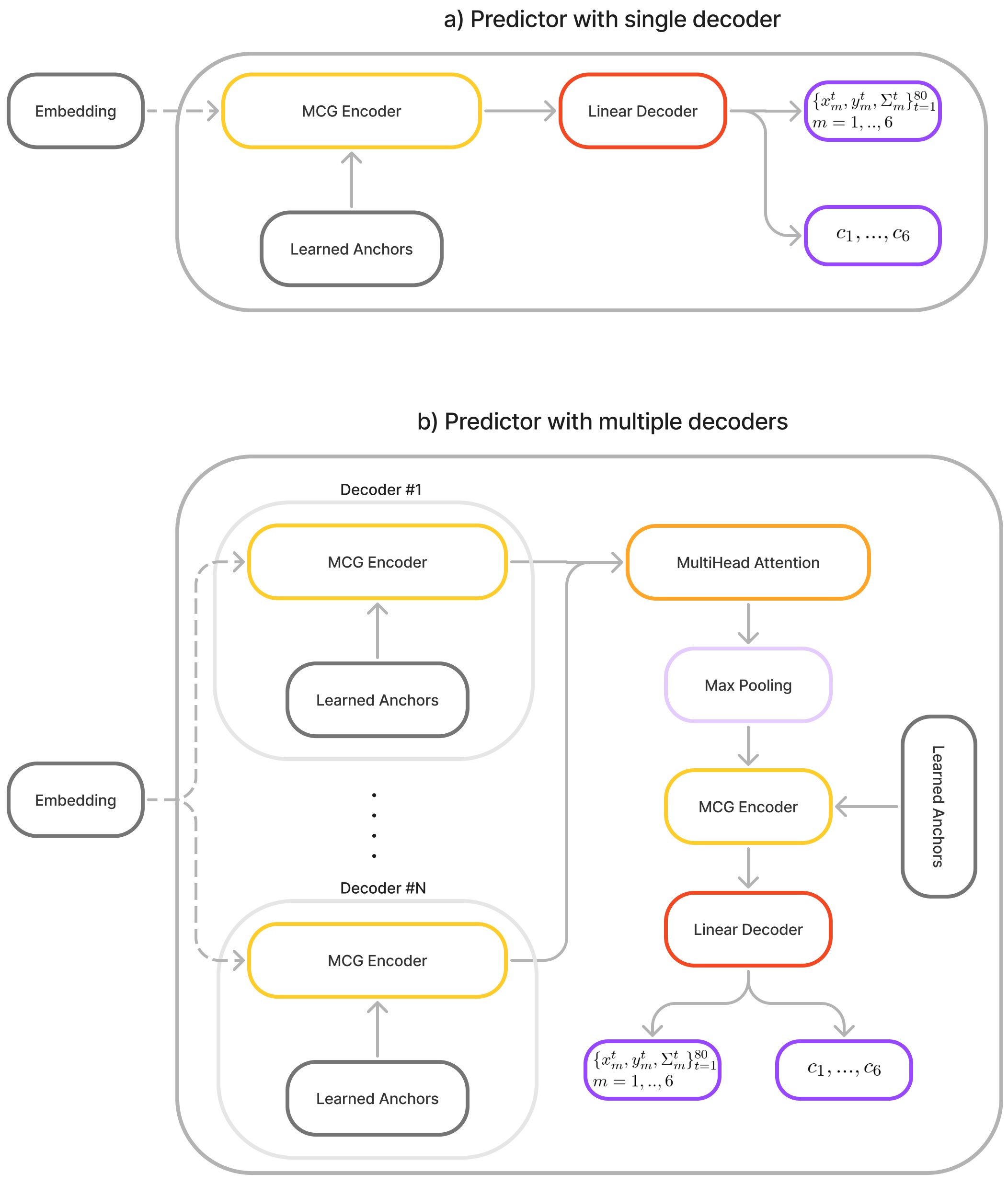}
\end{center}
   \caption{Architectures of the predictors. a) Predictor with a single decoder. b) Predictor with multiple decoders and attention \cite{attention} mechanism}
\label{fig:predictors}
\end{figure}

Our model follows the architecture of MultiPath++ \cite{mppp} mostly, however we tried to alternate it a little (see Fig.~\ref{fig:pipeline}). First we did not have a specific encoder for autonomous vehicle itself. Second we did not use proposed EM algorithm because in our experiments it performed numerically unstable. Instead we experimented with a single decoder with 6 modes and 5 decoders each with 6 modes followed by attention mechanism \cite{attention} and multi-context gating block (MCG) from MultiPath++ \cite{mppp} that mapped 30 modes into required 6. With multiple decoders we used a proposed strategy of blocking weighs update for randomly selected decoders (see Fig.~\ref{fig:predictors}).
\subsection{Training strategy}
For each target agent we predicted 6 modes, each consisted of 80 coordinates $\{(x^t_m, y^t_m)\}_{t=1}^{80}$ for $m=1,...,6$ along with covariance matrices $\{\Sigma^t_m\}_{t=1}^{80}$. For each mode we also predicted its probability $c_m$. Thus the likelihood has the form
$$
l = \sum_{m=1}^{6} c_m \prod_{t=1}^{80} N(\mu^t_{gt} - \mu^t_{m}, \Sigma^t_m)
$$
where $\mu^t_m = (x^t_m, y^t_m)$ predicted coordinates for timestamp $t$ and mode $m$ and $\mu^t_{gt} = (x^t_{gt}, y^t_{gt})$ is the corresponding ground truth coordinate. We used the negative log-likelihood as an objective function and optimized it w.r.t. the whole set of predicted trajectories.  In our experiments models with trained covariance matrices performed better. We trained our model for 1.5 million iterations with initial learning rate of $10^{-4}$ with Adam \cite{adam} optimizer and ReduceLROnPlateau scheduler.
\subsection{Augmentations}
While training the model we faced some overfitting especially for minor agent classes. The natural way to avoid this problem is to use augmentations, however is it not obvious what augmentations may provide a quality improvement in this specific task. Thus finally we have chosen to use masking for the history data. For each timestamp of the history we randomly put the values to zero and the validity flag to false during training. The probability of masking a single timestamp $p_{mask}=0.15$. We use masking only for historical data of both target and surrounding agents and do not mask the road graph.

\subsection{Postprocessing}
Since the proposed loss function does not directly optimize the target SoftMAP metric we decided to apply postprocessing. As MAP metric is typical for detection tasks we decided to use a typical approach - non-maximum suppression. This algorithm has been successfully used in multiple of previous works \cite{nms_dense_tnt,nms_hyper,nms_liu2021multimodal,nms_l2p,nms_motionnet}. More specifically, in case where two trajectories appear to be close enough to each other the less probable was suppressed in favour for more probable one. However as the number of input trajectories is equal to the number of the output ones we do not completely drop the suppressed trajectories but assign some minor constant probability to them.

\section{Results}

\begin{table}[h]
\centering
\begin{tabular}{l|ll}
Object Type & Soft mAP & mAP \\ \hline
Avg Vehicle & 0.4467 & 0.4382 \\
Avg Pedestrian & 0.3831 & 0.3758 \\
Avg Cyclist & 0.3493 & 0.3458 \\
Avg 3s & 0.4875 & 0.4773 \\
Avg 5s & 0.3935 & 0.3883 \\
Avg 8s & 0.2981 & 0.2943 \\
Total & 0.3930 & 0.3866
\end{tabular}
\caption{Detailed evaluation of our model on test set of Waymo Open Motion Dataset~\cite{waymo}.}
\label{tab:results_detailed}
\end{table}

For the final submission we selected the best model for each of agent types. For cars and pedestrians the model with single decoder performed the best, whereas for cyclist the model with multiple decoders followed by attention mechanism outperformed other models. Numerical results are presented in Tab.~\ref{tab:results_detailed}

{\small
\bibliographystyle{ieee_fullname}
\bibliography{egbib}
}

\end{document}